\documentclass[11pt]{article}
\usepackage{coling2018}
\usepackage[dvipsnames]{xcolor}
\usepackage{times}
\usepackage{microtype}
\usepackage{latexsym}
\usepackage{amsfonts}
\usepackage{amsmath}
\usepackage{mathrsfs}

\usepackage{url}
\usepackage{listings}
\usepackage{microtype}
\usepackage{array}
\usepackage{booktabs}
\usepackage{todonotes}
\usepackage{pgfplots}
\usepackage{linguex}
\usepackage{subcaption} 
\usepackage{caption}
\usepackage[inline]{enumitem} 
\usepackage[group-separator={,}]{siunitx}
\usepackage{tikz4nlp}
\pgfplotsset{major grid style={cb_gray!35}}

\title{Modeling Semantics with Gated Graph Neural Networks \\ for Knowledge Base Question Answering}

\author{Daniil Sorokin \and Iryna Gurevych\\
Ubiquitous Knowledge Processing Lab (UKP) \and Research Training Group AIPHES \\
Department of Computer Science, 
Technische Universit\"at Darmstadt \\
  {\url{www.informatik.tu-darmstadt.de/ukp/}} \\ 
}

\date{}

\begin{document}
\maketitle

\begin{abstract}
The most approaches to Knowledge Base Question Answering are based on semantic parsing. In this paper, we address the problem of learning vector representations for complex semantic parses that consist of multiple entities and relations.
Previous work largely focused on selecting the correct semantic relations for a question and disregarded the structure of the semantic parse: the connections between entities and the directions of the relations.
We propose to use Gated Graph Neural Networks to encode the graph structure of the semantic parse. 
We show on two data sets that the graph networks outperform all baseline models that do not explicitly model the structure. The error analysis confirms that our approach can successfully  process complex  semantic parses.

\end{abstract}

\section{Introduction}
\blfootnote{
    \hspace{-0.65cm}  
    This work is licensed under a Creative Commons 
    Attribution 4.0 International License.
    License details:
    \url{http://creativecommons.org/licenses/by/4.0/}
}

Knowledge base question answering (QA) is an important natural language processing problem. Given a natural language question, the task is to find a set of entities in a knowledge base (KB) that constitutes the answer. For example, for a question ``What is Princess Leia's home planet?'' the answer, ``Alderaan'', could be retrieved from a general-purpose KB, such as Wikidata\footnote{\url{https://www.wikidata.org/}}. A successful KB QA system would ultimately provide a universally accessible natural language interface to factual knowledge~\cite{Liang2016}.

QA requires precise modeling of the question semantics through the entities and relations available in the KB in order to retrieve the correct answer. Figure~\ref{fig:question-graph-simple} shows how the above example question could be modeled using Wikidata. The depicted graph structure consists of entities and relations from the KB and the special $q$-node. Any entity in the KB that can take the place of the $q$-node will be a part of the answer.

In this paper, we describe a semantic parsing approach to the problem of KB QA. That is, for each input question, we construct an explicit structural semantic parse (\textit{semantic graph}), as in Figure~\ref{fig:question-graph-simple}. Semantic parses can be deterministically converted to a query to extract the answers from the KB. Similar graphical representations were used in previous work~\cite{Yih2015,Reddy2016,Bao2016}.

However, the modern semantic parsing approaches usually focus either on the syntactic analysis of the input question~\cite{Reddy2016} or on detecting individual KB relations~\cite{Yu2017}, whereas \textit{the structure of the semantic parse} is ignored or only approximately modeled. \newcite{Reddy2016} use the syntactic structure of the question to build all possible semantic parses and then apply a linear model with manually defined features to choose the correct parse. A subset of their features encodes basic information about the graph structure of the semantic parse (e.g. number of nodes). The state-of-the-art approach of \newcite{Yih2015} and \newcite{Bao2016} restricts all semantic parses to a single core relation and a small set of constraints that can be added to it. Their system uses manual features for the constraints and a similarity score between the core relation and the question to model the semantic parses.

\begin{figure}[ht]
  \centering
  \begin{minipage}{0.37\linewidth}
  \begin{center}
    \begin{tikzpicture}
      \node (qvar) [graph_variable] {$q$};
      \node (entity1) [above=1cm of qvar,xshift=2cm,anchor=west,graph_entity] {Princess Leia};
      \node (entity2) [below=1cm of qvar,xshift=2cm,anchor=west,graph_entity] {planet};
      \draw[<-,rel_arrow] (qvar) -- node[text_above_arrow,midway] {home world} (entity1.west);
      \draw[->,rel_arrow] (qvar) -- node[text_above_arrow,midway] {instance of} (entity2.west);
    \end{tikzpicture}
  \end{center}
  \captionof{figure}{Semantic graph for an example question ``What is Princess Leia's home planet?''\label{fig:question-graph-simple}}
  \end{minipage}\hfill%
  \begin{minipage}{0.6\linewidth}
    \begin{center}
      \begin{tikzpicture}
      \begin{axis}[
          height=5cm,
          width=5.5cm,    
          ymax=0.65, ymin=0.25,
          enlarge y limits=0.1,
          enlarge x limits=0.25,
          reverse legend,
          legend style={at={(1.05,1.0)},draw=cb_gray,
              anchor=north west,legend columns=1,font=\small},         
            bar width=2ex,
            xtick=data,
           ylabel={F-score},
           ylabel near ticks,
           y label style={font=\small},
           ytick distance=0.1,
           tick label style={font=\small},    
           ymajorgrids     
            ] 
        \addplot[color=cb_purple,mark=triangle*] coordinates {(1, 0.45905782)
         (2, 0.39911309)
         (3, 0.40603028)
         (4, 0.2548137)
        };  
        \addplot[color=cb_darkblue,mark=triangle*] coordinates {(1, 0.58583996)
         (2, 0.45288994)
         (3, 0.45847174)
         (4, 0.29256626)
        };       
        \addplot[color=cb_lightblue,mark=*] coordinates {(1,  0.6191708)
         (2, 0.55988556)
         (3, 0.50304471)
         (4, 0.38990669)
        };  
        \addplot[color=cb_darkgreen,mark=square*,mark size=2] coordinates {(1, 0.60968578)
         (2, 0.54855154)
         (3, 0.50594326)
         (4, 0.38990669)
        };        
        \legend{\newcite{Berant2014},\newcite{Reddy2016}, \newcite{Jain},\newcite{Yih2015}}
        \end{axis}
        \end{tikzpicture}	
      \end{center}
      \captionof{figure}{F-score QA results from previous work by the number of relations needed to find the correct answer\label{fig:previous-work-results}}
  \end{minipage}
  \end{figure} 

The abovementioned systems were evaluated on the WebQuestions data set~\cite{Berant2013}. Figure~\ref{fig:previous-work-results} plots results for the state-of-the-art systems by the number of relations that needs to be identified to get the correct answer to a question.\footnote{We include results for the most recent systems that have published the output on the WebQuestions data set. We compute the number of needed relations from the manual semantic parses provided by \newcite{Yih2016}.} For example, the question in Figure~\ref{fig:question-graph-simple} requires two relations to find the answer. It can be clearly seen in Figure~\ref{fig:previous-work-results} that the lack of structure modeling in the modern approaches results in a worse performance on more complex questions that require more than one relation.

We claim that one needs to \textit{explicitly model the semantic structure} to be able to find the correct semantic parse for complex questions. In this paper, we address this gap and investigate ways to encode the structure of a semantic parse and to improve the performance for more complex questions. In particular, we adapt Gated Graph Neural Networks (GGNNs), described in \newcite{Li2015b}, to process and score semantic parses. To verify that GGNNs indeed offer an improvement, we construct a set of baselines based on the previous work that we train and evaluate in the same controlled QA environment. Throughout the experiments, we use the Wikidata  open-domain KB~\cite{Vrandecic2014} to construct semantic parses and retrieve the answers.\footnote{ At the moment, Wikidata contains more than 40 million entities and 350 million relation instances: \\ \raggedright \url{https://www.wikidata.org/wiki/Special:Statistics}}

\paragraph{Contributions} To summarize, the main contributions of our work are:
  \begin{enumerate}[label=(\roman*)]
    \item Our analysis shows that the current solutions for KB QA do not perform well on complex questions;
    \item We apply Gated Graph Neural Networks on directed graphs with labeled edges and adapt them to handle a large set of possible entities and relations types from the KB. To the best of our knowledge, we are the first to use GGNNs for semantic parsing and KB QA;
    \item Our Gated Graph Neural Network implementation for semantic parsing improves performance on complex questions in comparison to strong baselines. The results show a 27.4\% improvement of the F-score against the best non-graph model.
  \end{enumerate}

\paragraph{Code and data sets} 
Our system can be used with a pre-trained model to answer factual questions with Wikidata or trained anew on any data set that has question-answer pairs.
The complete code, the scripts that produce the evaluation data and the installation instructions can be found here: \url{https://github.com/UKPLab/coling2018-graph-neural-networks-question-answering}.

\section{Semantic parsing}

\subsection{Semantic graphs}

We employ structural semantic representations in the form of graphs to encode the meaning of a question. 
Our \textit{semantic graphs} consist of a question variable node ($q$), Wikidata entities ({\small\sffamily Taylor Swift}), relation types from Wikidata (\textsc{performer}) and constraints (see Figure~\ref{fig:question-graph-constraint} for an example graph with a constraint). The $q$-node is always present and denotes the answer to the question. That is, all entities from the KB that can take its place so that all relations and constraints hold, constitute the answer to the question. 

We write down a graph as a set of edges $\mathcal{E}$. Each edge links the $q$-node to one or two Wikidata entities (binary or ternary edge). The edge set $\mathcal{E}$ is defined via Wikidata relations and entities. Formally, Wikidata can be described as a very large graph $W = (E, R, I)$, where $E$ is a set of entities, $R$ is a set of binary relation types and $I$ is a collection of relation instances encoded as $r(e_1, e_2), \hspace{0.5em}r \in R, \hspace{0.5em}e_1,e_2 \in E$ (e.~g. \textsc{capital}\hspace{0.1em}({\small\sffamily Hawaii}, {\small\sffamily Honolulu})). Ternary relation instances in Wikidata are stored as a main relation triple and an attached modifer which itself is also a binary relation: $r_2\big(r_1(e_1, e_2), e_3\big), \hspace{0.5em}r_1, r_2 \in R, \hspace{0.5em}e_1,e_2,e_3 \in E$ (e.g. \textsc{character role}\hspace{0.1em}$\big($\textsc{cast member}\hspace{0.1em}({\small\sffamily Star Wars}, {\small\sffamily Carrie Fisher}), {\small\sffamily Princess Leia}$\big)$. Then, the edge set $\mathcal{E}$ of a semantic graph consists of binary and ternary edges that correspond to Wikidata relation instances with $q$ in place of one of the relation arguments. For temporal relations, the second argument can be either an $\mathrm{argmax}$ or an $\mathrm{argmin}$ constraint. This means that only entities that have the maximum or the minimum value of that relation are considered for the answer.
This definition represents a subset of first-order logic semantic parses restricted to conjunctions of predicates. The graphs are also isomorphic to SPARQL queries that can be used to evaluate a graph against the KB.

\begin{figure}[t]
  \begin{minipage}{0.4\linewidth}
  \begin{center}
    \begin{tikzpicture}
      \node (qvar) [graph_variable] {$q$};
      \node (entity1) [right=2.5cm of qvar,yshift=0.4cm,anchor=west,graph_entity] {Taylor Swift};
      \node (entity2) [right=2.5cm of qvar,yshift=-1cm,anchor=west,graph_entity] {album};
      \node (entity3) [below=of qvar,anchor=north,graph_modifier] {argmin};
      \draw[<-,rel_arrow] (qvar) -- node[text_above_arrow,midway] {performer} (entity1.west);
      \draw[->,rel_arrow] (qvar) -- node[text_above_arrow,midway] {instance of} (entity2.west);
      \draw[<-,rel_arrow] (qvar) -- node[text_above_arrow,midway,yshift=-0.2cm] {publication date} (entity3.north);   
    \end{tikzpicture}
  \end{center}
  \captionof{figure}{A semantic graph for an example question ``What was the first Taylor Swift album?''\label{fig:question-graph-constraint}}
  \end{minipage}\hfill%
  \begin{minipage}{0.57\linewidth}
    \begin{center}
      \begin{tikzpicture}
        \node (qvar) [graph_variable] {$q$};
        \draw[draw=black, dotted] (qvar)+(-3ex,-3ex) rectangle ++(3ex,3ex);
        \node [above=0.25cm of qvar] {$S^1$};

        \node (qvar1) [graph_variable, right=1.5cm of qvar,yshift=1.15cm] {$q$};        
        \node (qvar2) [graph_variable, right=1.5cm of qvar] {$q$};
        \node (qvar3) [graph_variable, right=1.5cm of qvar,yshift=-1.15cm] {$q$};
        \draw[->,rel_arrow,dashed,shorten >=1.5ex,shorten <=1.5ex] (qvar) -- node[midway,yshift=2ex] {$a_e$} (qvar1.west);
        \draw[->,rel_arrow,dashed,shorten >=1.5ex,shorten <=1.5ex] (qvar) -- node[midway,yshift=2ex] {$a_e$} (qvar2.west);
        \draw[->,rel_arrow,dashed,shorten >=1.5ex,shorten <=1.5ex] (qvar) -- node[midway,yshift=2ex] {$a_e$} (qvar3.west);
        \node (entity1) [right=3cm of qvar1,anchor=west,graph_entity] {Taylor Swift};
        \node (entity2) [right=3cm of qvar2,anchor=west,graph_entity] {Taylor Swift};
        \node (entity3) [right=3cm of qvar3,anchor=west,graph_entity] {Taylor Swift};
        \draw[->,rel_arrow] (qvar1) -- node[text_above_arrow,midway] {performer} (entity1.west);
        \draw[->,rel_arrow] (qvar2) -- node[text_above_arrow,midway] {has part} (entity2.west);
        \draw[->,rel_arrow] (qvar3) -- node[text_above_arrow,midway] {influenced} (entity3.west);
        
        \draw[draw=black, dotted] (qvar1)+(-3ex,-3ex) rectangle ++(5.5cm,3ex);
        \node [right=5.5cm of qvar1.north,yshift=-0.2cm] {$S^2$};
        \draw[draw=black, dotted] (qvar2)+(-3ex,-3ex) rectangle ++(5.5cm,3ex);
        \node [right=5.5cm of qvar2.north,yshift=-0.2cm] {$S^3$};
        \draw[draw=black, dotted] (qvar3)+(-3ex,-3ex) rectangle ++(5.5cm,3ex);
        \node [right=5.5cm of qvar3.north,yshift=-0.2cm] {$S^4$};
        
      \end{tikzpicture}
    \end{center}
    \captionof{figure}{A single graph generation step: applying the add entity action $a_e$ on an empty graph\label{fig:question-graph-generation}}    
  \end{minipage}
\end{figure}

Our semantic graphs were inspired by the query graphs of \newcite{Yih2015} and their extension in  \newcite{Bao2016}, the key difference being that we do not differentiate between the core relation and modifiers, but rather allow graphs that have multiple independent relations. We also allow relations to attach to other nodes rather than the $q$-node, enabling longer relation paths between a known entity and the $q$-node. Thus, the semantic graphs defined here are a superset of the query graphs in \newcite{Yih2015} and allow us to model more complex meanings.
Consequently, they also correspond to the formalized representations used by \newcite{Reddy2014} and the simple $\lambda$-DCS~\cite{Berant2013}, since those were the foundation for the query graphs~\cite{Yih2015}.

\subsection{Semantic graph construction}
\label{sec:graph-construction}

\newcite{Yih2015} defined a step-by-step staged graph generation that does not need full syntactic parses and was also adopted in subsequent publications \cite{Bao2016,Peng2017}. We use the same procedure as \newcite{Yih2015} to construct semantic graphs with a set of modifications that allow us to build more expressive and complex graphs.

We define a set of states and a set of actions that can be applied at each state to extend the current semantic graph. A state is a tuple of a graph and a set of free entities: $\mathcal{S} = (\mathcal{E}, \mathcal{F})$, where the graph is $\mathcal{E}$, as defined above, and $\mathcal{F} = \{e | e \in E \}$. The set of free entities $\mathcal{F}$ is the entities that were identified in the question but were not yet added to the graph.\footnote{We use an off-the-shelf entity linker to identify and disambiguate entity mentions in the question (see Section~\ref{sec:inference}).} The first state is a tuple of an empty graph with no edges and a set of all entities identified in the question $\mathcal{S}^1 = (\{\}, \mathcal{F})$.

Let $\mathcal{A} = \{a_e, a_c, a_m\}$ be the set of actions that can be taken to extend the graph at the current state. The action $a_e$ (\textit{add entity}) pops a free entity $e$ from the set $\mathcal{F}$ at the current state $\mathcal{S}^t$. Then it queries the KB and retrieves the set of available relation types $R_e$ for the entity $e$. For each relation type $r \in R_e$, it creates a new copy of the graph and adds a new directed edge between the $q$-node and $e$ with the relation type $r$:  $a_e(\mathcal{E}, \mathcal{F})= 
\{ \mathcal{E} \cup r(q,e), \mathcal{F} - e \hspace{0.5em}|\hspace{0.5em}e\in \mathcal{F}, r \in R_e\}$. Contrary to \newcite{Yih2015} and \newcite{Bao2016}, we allow two-step paths between the $q$-node and the entity $e$: $r(q,d) \land r(d, e)$.\footnote{Freebase relation instances always have an intermediate node and a two-step path corresponds to a single step in Wikidata.}

The action $a_c$ (\textit{add constraint}) pops a free entity $e$ and follows the same procedure as $a_e$, but instead of adding a new edge, it adds a modifier to the last added edge of the graph, thus creating a ternary edge: $a_c(\mathcal{E}, \mathcal{F})= 
\{ \mathcal{E} \cup r_2(r_1(q, e_1), e_2), \mathcal{F} - e_2\hspace{0.5em}|\hspace{0.5em}e_2\in \mathcal{F}, r_2 \in R_{e_2}, r_1(q, e_1) \in \mathcal{E} \}$. Finally, the action $a_m$ (\textit{add argmax/argmin}) adds a new edge with either $\mathrm{argmax}$ or $\mathrm{argmin}$ sorting constraint to the semantic graph: $a_m(\mathcal{E}, \mathcal{F})= 
\{ \mathcal{E} \cup r(q, \mathrm{argmax}), \mathcal{F}; \mathcal{E} \cup r(q, \mathrm{argmin}), \mathcal{F}\hspace{0.5em}|\hspace{0.5em}r \in R_d \}$, where $R_d$ is a set of KB relation types that allow dates as values. Our semantic graph construction process allows to effectively search the space of possible graphs for a given question through an iterative application of the defined actions $\mathcal{A}$ on the last state $\mathcal{S}^{t}$ (see, for example, Figure~\ref{fig:question-graph-generation}).

\section{Representation learning}

We follow the state-of-the-art approaches for QA~\cite{Yih2015,Dong2015,Bao2016} and learn representations for the question and every possible semantic graph. Then we use a simple reward function $\gamma$ to judge if a semantic graph is a correct semantic parse of the input question. Below we describe the architectures that we use to learn the representations for questions and graphs. 

\subsection{Deep Convolutional Networks}
\label{sec:dcnn}

\begin{figure}[t]
  \begin{minipage}{0.49\linewidth}
\begin{center}
\includegraphics[width=0.99\linewidth]{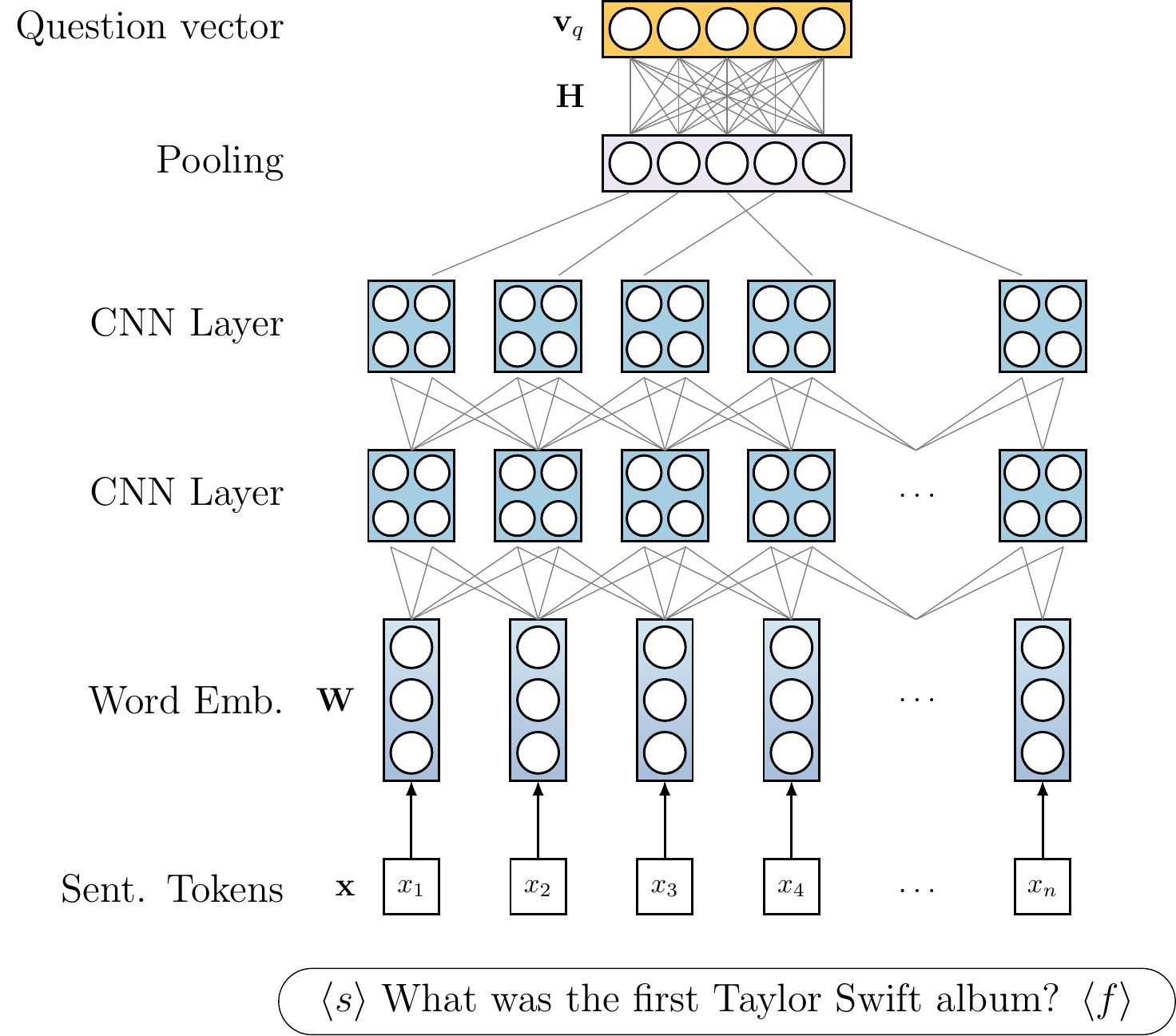}
\end{center}
\captionof{figure}{Deep Convolutional Neural Networks (DCNN) architecture, here used to process an example question\label{fig:question-encoder-architecture}}
\end{minipage}\hfill%
\begin{minipage}{0.49\linewidth}
   \begin{center}
    \includegraphics[width=0.98\linewidth]{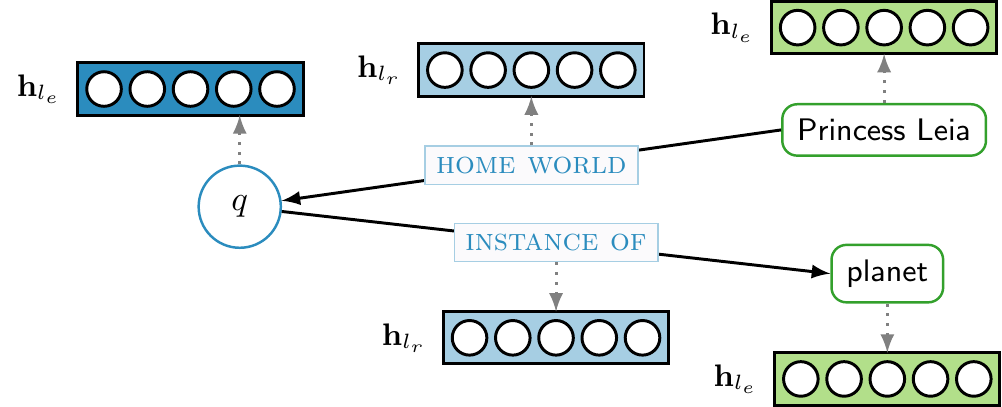}
    \end{center}
    \captionof{figure}{Encoding a graph into initial hidden states\label{fig:gnn-encoder-architecture}}
    \begin{center}
      \includegraphics[width=0.7\linewidth]{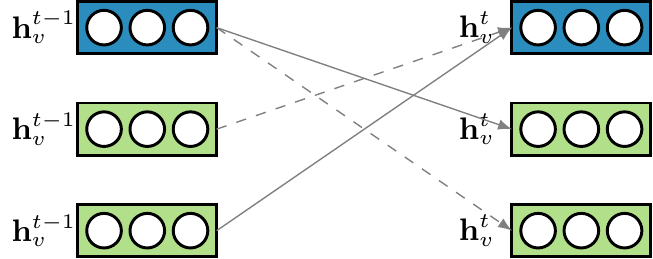}
      \end{center}
      \captionof{figure}{Unrolled recurrence for one timestep\label{fig:gnn-recurrence}, solid lines show updates along the direction of the relation in the graph and the dashed lines in the opposite direction}
  \end{minipage}
\end{figure}

We use Deep Convolutional Neural Networks (DCNN) to learn a representation for a question. DCNNs have been proven effective for constructing sentence-level representations on a variety of NLP tasks, including named entity recognition~\cite{Strubell2017} and KB QA~\cite{Yih2015,Dong2015}.  

The DCNN architecture is depicted in Figure~\ref{fig:question-encoder-architecture}, where it is used to map an input question to a fixed-size vector representation. The input question is first tokenized and the special start and end tokens are added to the sequence: $\mathbf{x} = \{\langle s \rangle, x_1, x_2 \ldots x_n, \langle f \rangle\}$.
Next, we map the tokens in $\mathbf{x}$ to $d_w$-dimensional pre-trained word embeddings, using a matrix $\mathbf{W_{\mathit{glove}}} \in \mathbb{R}^{|V_w| \times d_w}$, where $|V_w|$ is the size of the vocabulary. We use 50-dimensional GloVe embeddings that were trained on a 6 billion corpus~\cite{Pennington2014}. 
The sequence of word embeddings is further processed by an array of CNN layers. 
We apply a pooling operation after the last CNN layer and transform the output with a fully connected layer $\mathbf{H}$ and a $\mathrm{ReLU}$ non-linearity. We take the resulting vector $\mathbf{v}_q$ as the representation of the question.

\subsection{Gated Graph Neural Networks}
\label{sec:ggnn}

Gate Graph Neural Networks (GGNN) process graphs by iteratively updating representations of graph nodes based on the neighboring nodes and relations~\cite{Li2015b}.
We adopt GGNNs for semantic parsing to learn a vector representation of a semantic graph. \newcite{Li2015b} give a formulation of GGNNs for graphs with labeled nodes and typed directed edges. We extend their formulation to include labeled edges. To the best of our knowledge, we are the first to apply GGNN to semantic parsing and KB QA.

For a graph $\mathcal{E}$, we extract a set of all entities $\mathcal{V}$ in the graph and a set of all relation types $\mathcal{R}$ of its edges. We use labels from Wikidata to compute vectors for entities and relation types (Figure~\ref{fig:gnn-encoder-architecture}). This enables us to directly incorporate the information on millions of entities and hundreds of relation types from the KB.
For an entity $e \in \mathcal{V}$ or a relation type $r \in \mathcal{R}$ we retrieve the label and tokenize it: $\mathbf{l} = \{l_1, l_2 \ldots l_n\}$. Then we map each token in $\mathbf{l}_e$, $\mathbf{l}_r$  to a word embedding using the matrix $\mathbf{W_{\mathit{glove}}}$, sum them and process with a fully connected layer to get a single label vector: $\mathbf{h}_l = \tanh \big(\mathbf{W_l}\big[\sum_{n=1}^{|\mathbf{l}|} w_n \big] + \mathbf{b_l}  \big)$. We initialize the hidden states for the graph nodes with the label vectors of the entities: $\mathbf{h}_v^{(1)} = \mathbf{h}_{l_e}$. We further transform the relation label vectors to get directional embeddings for relations types: $\mathbf{h}_r' = \mathbf{W_{\rightarrow}} \mathbf{h}_{l_r}$, $\mathbf{h}_r'' = \mathbf{W_{\leftarrow}} \mathbf{h}_{l_r}$. Using the same word embeddings as an input to construct the question and relation representations has been shown successful in previous work~\cite{Jain}.

\paragraph{Propagation Model} GGNNs are a type of recurrent neural networks. The recurrence is unrolled for a fixed number of steps $T$ and the gating mechanism works akin to Gated Recurrent Units~\cite{Cho2014}. The propagation model for GGNN is defined as follows:

\begin{minipage}{0.44\linewidth}
\begin{align}
\begin{split}
  \mathbf{a}_v^{(t)} &= \mathbf{A}_{v:}^\top\big[\mathbf{h}_1^{(t-1)\top} \ldots  \mathbf{h}_{|\mathcal{V}|}^{(t-1)\top}\big] \\ &+ \mathbf{A'}_{r:}^\top\big[\mathbf{h}_1'^{\top} \ldots  \mathbf{h}_{|\mathcal{R}|}'^{\top}, \mathbf{h}_1''^{\top} \ldots  \mathbf{h}_{|\mathcal{R}|}''^{\top}\big]
  \end{split}
  \label{eq:ggnn-activations} \\[1ex]
  \mathbf{z}_v^t &= \sigma \big(\mathbf{W}^z\mathbf{a}_v^{(t)} + \mathbf{U}^z\mathbf{h}_v^{(t-1)} + \mathbf{b}^z\big)
  \label{eq:ggnn-update}
\end{align}
\end{minipage}\hfill%
\begin{minipage}{0.5\linewidth}
\begin{align}
  \mathbf{r}_v^t &= \sigma \big(\mathbf{W}^r\mathbf{a}_v^{(t)} + \mathbf{U}^r\mathbf{h}_v^{(t-1)} + \mathbf{b}^r\big)
  \label{eq:ggnn-reset} \\[1ex]
  \widetilde{\mathbf{h}_v^{(t)}} &= \tanh \big(\mathbf{W}\mathbf{a}_v^{(t)} + \mathbf{U}\big(\mathbf{r}_v^t\odot\mathbf{h}_v^{(t-1)}\big) + \mathbf{b}\big) \\[1ex]
  \mathbf{h}_v^{(t)} &= (1-\mathbf{z}_v^t) \odot \mathbf{h}_v^{t-1} + \mathbf{z}_v^t\odot\widetilde{\mathbf{h}_v^{(t)}},
\end{align}
\end{minipage}\vspace{2ex}
where $\sigma$ is the logistic sigmoid function and $\odot$ is the element-wise multiplication.
The matrix $\mathbf{A} \in \mathbb{R}^{|\mathcal{V}| \times 2|\mathcal{V}|}$ stores the structure of the graph: a row of the matrix $\mathbf{A}_{v:}$ records the edges between the node $v$ and the other nodes in the graph (we differentiate between incoming and outgoing edges). The second matrix $\mathbf{A}' \in \mathbb{R}^{|\mathcal{V}| \times 2|\mathcal{R}|}$ stores the relation types of the incoming and outgoing edges.

The main difference from the model defined in \newcite{Li2015b} is that we compute activations $\mathbf{a}_v^{(t)}$ based on both the node hidden vectors $\mathbf{h}_v^{(t-1)}$ and relation hidden vectors $\mathbf{h}'_r$ (Eq~\ref{eq:ggnn-activations}). The $\mathbf{z}_v^t$ in Eq~\ref{eq:ggnn-update} and $\mathbf{r}_v^t$ in Eq~\ref{eq:ggnn-reset} are update and reset gates, that incorporate information from nodes, relation types and from the previous step to update nodes' hidden states at each iteration. We do not make updates to the hidden vectors of the relations and use them only to pass the information to the nodes' hidden states.   

\paragraph{Graph-level Output Vector} We unroll the recurrence for $T=5$ steps in the experiments (Figure~\ref{fig:gnn-recurrence}). To produce a graph-level output vector, we take the hidden vector for the $q$-node at the last time step $t=T$ and transform it with a fully-connected layer and the $\mathrm{ReLU}$ non-linearity:
$ \mathbf{v}_g = \mathrm{ReLU} \big(\mathbf{W}\mathbf{h}_q^{(t)} + \mathbf{b} \big)$.

\section{Experiments}

\subsection{Data}
\label{sec:datasets}

We use Wikidata to show experimentally that GGNNs are better at learning representations of semantic graphs than previous approaches. Hence, we choose two data sets that can be processed with Wikidata to compare the GGNNs to other models: WebQSP-WD and QALD-7.

\paragraph{WebQSP-WD} We derive WebQSP-WD from WebQSP~\cite{Yih2016}, which is a corrected version of the popular WebQuestions data set~\cite{Berant2013}. WebQSP contains natural language questions, Freebase IDs of the correct answers and SPARQL queries to retrieve them that also use Freebase. Freebase was a common choice of a KB among the previous work, but was discontinued and is no longer up-to-date, including unavailability of APIs and new dumps. 
The questions in the data set were collected with the Google Suggest API and are thus more `natural' than manually constructed questions. The answers were retrieved from Freebase with the help of crowd-sourcing. The data set contains both simple questions that can be answered with a single relation as well as complex questions that require multiple relations and constraints. It is a common benchmark for semantic parsers and information retrieval systems and was used in the most recent studies on KB QA.
We automatically map Freebase IDs in the WebQSP train and test sets to Wikidata IDs and filter out questions which answers do not have the mapping.  We designate this version of the dataset WebQSP-WD. It is important to note that this does not ensure that a question is answerable with Wikidata as there still might be no relation paths in the KB that connect the entities in the question with the answer.

\paragraph{QALD-7}  As the second data set, we use QALD-7 that was developed for Task 4 of the QALD-7 Shared Task, ``English question answering over Wikidata''~\cite{10.1007/978-3-319-69146-6_6}. The QALD-7 data set contains a small number of manually constructed complex questions that were specifically created to test system's ability to process questions with multiple entities and constraints. The data set uses Wikidata IDs for all annotations. We do not train on QALD-7 data set and use it solely for an out-of-domain evaluation. The data set statistics can be found in Table~\ref{table:datasets}.

\subsection{Models}

We define five models for our experiments, including three baselines and two graph models. We use cosine similarity between the question representation and the semantic graph representation as a reward function that judges whether the semantic graph is the correct parse of the question: $\gamma = \mathrm{cos}(\mathbf{v_s}, \mathbf{v_g})$.

\begin{enumerate}
  \item STAGG (re-implementation of \newcite{Yih2015}) --- We implement a model that uses a combination of a neural network and manual features suggested in \newcite{Yih2015} and in the follow up work of \newcite{Bao2016}. The approach of \newcite{Yih2015} performed best among the previous work on complex questions (see Figure~\ref{fig:previous-work-results}). First, DCNN is used to produce a representation for the input question, as described in Section~\ref{sec:dcnn}. Then, we replace the entity tokens in the input question with a special entity symbol $\langle e \rangle$  and apply DCNN on it again to get a second representation for the question. For each semantic graph, we take the label of the relation in the first edge (\textit{core edge}), tokenize it and likewise apply DCNN on it to get a representation. We produce two representations for the core relation: one that includes the label of the attached entity and one that includes the entity symbol. The manual features include binary indicators for modifiers and constraints in the semantic graphs, as well as features for certain keywords in the question (see \newcite{Yih2015} for a detailed description). 
  
The original system and the models of \newcite{Yih2015} and of \newcite{Bao2016} are not available and therefore we use our own implementation of their approach. There is a number of small difference with the original model: we use a deep CNN instead of a single CNN layer and train the DCNN together with the manual features, whereas \newcite{Yih2015} pre-trained the CNN model on a separate corpus and used its output in the feature model.

  \item Single Edge model --- We use the DCNN to encode the question and the label of the first edge of a semantic graph. The rest of the information is ignored.
  \item Pooled Edges model --- We use the DCNN to encode the question and the label of each edge in the semantic graph. To get a fixed-vector representation of the graph, we apply a pooling operation over the representation of the individual edges. This model encodes all information about the semantic graphs, but disregards their structure.
  \item Graph Neural Network (GNN) --- To judge the effect of the gated graph neural architecture, we also include a model variant that does not use the gating mechanism and directly computes the hidden state as a combination of the activations (Eq~\ref{eq:ggnn-activations}) and the previous state.
  \item Gated Graph Neural Network (GGNN) --- We use the GGNN to process semantic parses, as described in Section~\ref{sec:ggnn}. This model encodes all information from semantic graphs, including their structure, into a vector representation. To encode the question we use the same DCNN model (see Section~\ref{sec:dcnn}).
    
\end{enumerate}
The defined baselines use either manual features to capture the structure of the semantic graph (STAGG), a simple pooling mechanism (Pooled Edges) or disregard the structure completely (Single Edge). The two graph models (GNN and GGNN) make the full use of the graph structure of a semantic parse and encode it with graph neural networks. With this set-up, we are able to demonstrate what effect different levels of inclusion of the graph structure into the model have on the final performance for KB QA. We do not include the published models of \newcite{Berant2013} and \newcite{Reddy2014} in the comparison since they were trained on Freebase and are not compatible with Wikidata. We use the more recent STAGG approach to position the graph models against the previous work and the feature-based models.

\subsection{Training the model}

To train the model, we need positive pairs of questions and semantic graphs. Since WebQSP does not contain semantic parses for Wikidata, we use weak supervision as suggested in~\newcite{Berant2013}. 
Specifically, we follow \newcite{Yih2015} and run our semantic graph construction procedure to create training instances (see Section~\ref{sec:graph-construction}). 
We use the state-of-the-art S-MART entity linking system for noisy data~\cite{Yang2015a} to extract a set of entities $\mathcal{F}$ from each question.\footnote{S-MART is not openly available, but the output on the WebQuestions dataset was made available by~\newcite{Yih2015}: \url{https://github.com/scottyih/STAGG}}
Instead of scoring the semantic graphs with the model, we evaluate each graph against Wikidata and compare the extracted answers to the manual answers in the data set. The semantic graphs that result in a correct answer are stored as positive training instances and the rest of the graphs generated during the same process are used as negative instances. 
Due to the differences between Freebase and Wikidata, some question can not be answered exactly using Wikidata.
We generate positive semantic graphs for 1945 questions out of 2880 (see Table~\ref{table:datasets}) and put 628 as a development set aside.

\textbf{Practical considerations}
At each training epoch, we take all positive semantic graphs and up to 100 negative graphs per question. We optimize the maximum margin loss function: $\mathcal{L} = \sum_{g\in C} \max\big(0, (m - \gamma(\mathbf{v}_q,\mathbf{v}_g^+) + \gamma(\mathbf{v}_q,\mathbf{v}_g^-))\big)$, where $C$ is a set of semantic parses for the given question. From the loss function, we compute updates for the GGNN and the DCNN parts of the model.

All models are trained using the Adam optimizer \cite{Kingma2014a} with a batch size of $64$. We use an early stopping criterion on the development data to determine the number of training epochs. 
The learning rate is fixed to $0.001$ and the other optimization parameters are set as recommended in \newcite{Kingma2014a}:  $\beta_1=0.9$, \mbox{$\beta_2=0.999$}, $\epsilon=1e-08$. We apply Dropout~\cite{Srivastava2014} at every fully-connected layer as well as on the embeddings layers.
We determine the hyper-parameters, such as the size of the hidden layer, with the random search on the development set (see Table~\ref{table:hyper-params}). On the 1945 training questions, the GGNN model has usually finished training in under two hours on a single GPU.

\begin{figure}[!ht]
  \begin{minipage}{0.49\linewidth}  
  \begin{center}
  \begin{tabular}{>{\raggedright}p{0.37\linewidth}
  >{\raggedleft}p{0.11\linewidth}
  >{\raggedleft}p{0.11\linewidth}
  >{\raggedleft\arraybackslash}p{0.185\linewidth}}
  \toprule 
    &  \multicolumn{2}{c}{WebQSP-WD} & QALD-7 \\  
    & (train) & (test) &   \\
  \midrule
questions  & \num{2880} & \num{1033} & \num{80}\\
complex questions & \num{419} & \num{293} & \num{42}\\
  avg. \# of relations per question & \num[round-precision=2]{1.354422207876049} & \num[round-precision=2]{1.3872216844143272} & \num[round-precision=2]{2.125} \\
  \bottomrule
  \end{tabular} 
  \end{center}
  \captionof{table}{Dataset statistics \label{table:datasets}}
\end{minipage}\hspace{0.5ex}%
\begin{minipage}{0.51\linewidth}  
  \begin{center}
    \begin{tabular}{p{0.34\linewidth}
    >{\raggedleft}p{0.3\linewidth}
    >{\raggedleft\arraybackslash}p{0.17\linewidth}}
    \toprule 
    Parameter & Tested values & Selected\\ 
    \midrule
    Hidden layer size & $[32, 64, 128, 256, 512]$ & $256$\\ 
    CNN filter size & $[32, 64, 128, 256, 512]$ & $256$\\ 
    Dropout ratio & $\mathcal{U}(0.0, 0.5)$ & $0.2$\\
    \# of CNN layers & $[1, 2, 3]$ & $2$\\
    Pooling operation & $[\mathrm{avg},\mathrm{sum}, \mathrm{max}]$ & $\mathrm{max}$\\
    \bottomrule
    \end{tabular}
    \end{center}
    \captionof{table}{Optimized hyper-parameter values \label{table:hyper-params}}  
\end{minipage}
\end{figure}

\subsection{Inference}
\label{sec:inference}

We take the steps described in Section~\ref{sec:graph-construction} to construct possible semantic graphs for a question at inference time. For WebQSP-WD, we use the entities produced by S-MART~\cite{Yang2015a} to start the graph construction. On QALD-7, we use the annotated entities provided by the Shared Task organizers.

Each question and semantic graph are encoded into fixed-size vector representations and the reward function $\gamma$ is used to score the graphs. The highest scoring graph is used to retrieve the answers from Wikidata. Given the iterative nature of our semantic graph construction procedure, we adopt beam search to speed up the computation. We score the graphs after each step and select the top $10$ to proceed.

\section{Results}
\label{sec:results}

We follow the previous work~\cite{Berant2013} and use precision, recall and F-score to evaluate the models. The measures are computed for each individual question and then macro-averaged. This ensures a fair evaluation, since a system might provide a partially correct answer that is nevertheless better than a complete miss. We compare the graph models to the baselines including the previous state-of-the-art STAGG architecture.\footnote{Since we use Wikidata and WebQSP-WD, the values reported in this work are not directly comparable to those for Freebase.}

\begin{figure}[t]
  \begin{minipage}{0.49\linewidth}   
  \begin{center}
  \begin{tabular}{p{0.275\linewidth}
  >{\raggedleft}p{0.15\linewidth}
  >{\raggedleft}p{0.125\linewidth}      
  >{\raggedleft\arraybackslash}p{0.125\linewidth}}
  \toprule 
  & P & R &  F \\
  \midrule
  STAGG  & \num{0.19105924} & \num{0.22672522} & \num{0.18276838} \\   \midrule  
  Single Edge  & \num{0.22397556} & \num{0.27129068} & \num{0.21476376} \\   Pooled Edges   & \num{0.20943364} & \num{0.25532975} & \num{0.20321495}\\   GNN   & \num{0.24185018} & \num{0.28901665} & \num{0.23259515}\\      \textbf{GGNN}   & \textbf{\num{0.26857575}} & \textbf{\num{0.31791255}} & \textbf{\num{0.25877491}}\\
  \bottomrule
  \end{tabular} 
  \end{center}
  \captionof{table}{Results on the WebQSP-WD test set \label{table:results-webqsp-wd}}
  \end{minipage}\hfill%
  \begin{minipage}{0.49\linewidth}  
    \begin{center}
      \begin{tabular}{p{0.275\linewidth}
        >{\raggedleft}p{0.15\linewidth}
        >{\raggedleft}p{0.125\linewidth}      
        >{\raggedleft\arraybackslash}p{0.125\linewidth}}
    \toprule 
    & P & R &  F \\
    \midrule
    STAGG &  \num{0.19336654} & \num{0.24626118} & \num{0.18612448} \\
    \midrule  
    Single Edge  & \num{0.21725} & \num{0.19627099} & \num{0.1951257} \\
    Pooled Edges   & \num{0.19036458} & \num{0.17998367} & \num{0.1605478}\\
    GNN & \num{0.1652105} & \num{0.20716295} & \num{0.17027307}\\
    \textbf{GGNN}  & \textbf{\num{0.21759998}} & \textbf{\num{0.27510635}} & \textbf{\num{0.21309655}}\\
    \bottomrule
    \end{tabular} 
    \end{center}
    \captionof{table}{Results on the QALD-7 data set \label{table:qald-results}}    
  \end{minipage}
  \end{figure}

\subsection{WebQSP-WD} 

We compare the results on the WebQSP-WD data set in Table~\ref{table:results-webqsp-wd}. As can be seen, the graph models outperform all other models across precision, recall and F-score, with GGNN showing the best overall result. We confirm thereby that the architecture that encodes the structure of a semantic parse has an advantage over other approaches. To validate the results, we have re-trained the model with different random seeds and observed little variance in the results (F-score, $\sigma=\num{0.02045640670855629}$).

The STAGG architecture delivers the worst results in our experiments, the main reason being supposedly that the model had to rely on manually defined features that are less flexible. The Single Edge model outperforms the more complex Pooled Edges model by a noticeable margin. The Single Edge baseline prefers simple graphs that consist of a single edge which is a good strategy to achieve higher recall values. 

Since our main goal is to produce better encoding of semantic graphs, we break down the performance of the evaluated models by the number of relations that are needed to find the correct answer.\footnote{We use again the manually constructed queries provided by~\newcite{Yih2016} to estimate it.} In Figure~\ref{fig:results-by-edge}, we see that for the STAGG and Single Edge baselines the performance on more complex questions drops compared to the results on simpler questions. The Pooled Edges model maintains a better performance across questions of different complexity, which shows the benefits of encoding all graph edges. 
Looking at Figure~\ref{fig:results-by-edge} we also get a further insight into the difference between the Single Edge and the Pooled Edges models. These two models achieve almost identical results on the simple questions, which is to be expected since these models are equivalent when the number of edges is $1$. On other questions, the Single Edge baseline performs mostly under the Pooled Edges model, but significantly outperforms it on the questions that need $4$ edges to get the correct answer.

We see that the GGNN model offers the best results both on simple and complex questions, as it effectively encodes the structure of semantic graphs. The performance of the model drops for the most complex questions (\# of edges $\geq 4$). That does not happen for the GNN model variant without the gating mechanism. We conjecture that this happens because GGNN has more parameters than GNN and therefore needs more data to learn. By looking at the model errors on the most complex questions, we could see that GGNN tends to incorrectly predict one of the relations in the graph, which results in a wrong answer. GNN, on the other hand, more often predicts less relations that are needed and therefore gets a non-zero score for a partially correct answer. For example for the question ``Who is the prime minister of Spain 2011?'', GNN predicts a graph of two relations \textsc{instance of}\hspace{0.1em}({\small\sffamily human}) and \textsc{head of government} which returns a list of all Spanish prime ministers. The complete correct graph would also include temporal constraints.

Notably, GGNN also has the best performance on the most simple semantic parses that only have one edge. In these cases, two nodes in the graph interact with each other through the single edge in both directions. The GGNN is better at capturing this interaction than other models.

\begin{figure}[t]
  \begin{minipage}{0.55\linewidth}   
    \begin{center}   
      \begin{tikzpicture}
      \begin{axis}[
          height=6.5cm,
          width=6.5cm,    
          ymax=0.35, ymin=0.0,
          ytick distance=0.1,
          enlarge y limits=0.1,
          enlarge x limits=0.25,
          reverse legend,
            legend style={at={(1.05,1.0)},draw=cb_gray,
              anchor=north west,legend columns=1,font=\small},         
            bar width=2ex,
         xtick=data,
         ylabel={F-score},
         ylabel near ticks,
         y label style={font=\small},
         ymajorgrids,
           tick label style={font=\small},         
            ]  
        \addplot[color=cb_darkyellow,mark=o,mark size=2] coordinates {(1, 0.17130449)
        (2, 0.24896035)
        (3, 0.11337868)
        (4, 0.10180624)
        };   
        \addplot[color=cb_purple,mark=*,mark size=2] coordinates {(1, 0.20125786)
        (2, 0.27519564)
        (3, 0.17988044)
        (4, 0.17030846)
        };            
        \addplot[color=cb_darkblue,mark=square,mark size=2] coordinates {(1, 0.17726405)
        (2, 0.29244968)
        (3, 0.24902651)
        (4, 0.12643678)
        };      
        \addplot[color=cb_darkgreen,mark=triangle*] coordinates {(1, 0.20817466)
         (2, 0.31443089)
         (3, 0.26433136)
         (4, 0.1954023)
        };           
        \addplot[color=cb_tudred,mark=square*] coordinates {(1, 0.2497884)
         (2, 0.31263496)
         (3, 0.27730365)
         (4, 0.05747126)
        };      
        \legend{STAGG, Single Edge, Pooled, GNN, GGNN}
        \end{axis}
\begin{axis}[
            ybar,
          height=3.75cm,
          width=6.5cm,   
          hide x axis, 
            enlarge y limits=0.05,
           enlarge x limits=0.25,
          ymax=800, ymin=0,
            bar width=2ex,
            ylabel={\# of questions},
            yticklabel pos=right,  
            ytick style={draw=none},  
            y tick label style={font=\footnotesize, cb_yellow},
            ylabel near ticks,  
            y label style={font=\small},
            ytick distance=250,       
            ]
        \addplot[cb_yellow,fill=cb_yellow!20] coordinates {(1, 740)
         (2, 215)
         (3, 49)
         (4, 29)
        };
        \end{axis}          
        \end{tikzpicture}	
      \end{center}
      \captionof{figure}{Evaluation results by \# of relations needed to find the correct answer (bars show \# of questions)
      \label{fig:results-by-edge}}
\end{minipage}\hfill%
\begin{minipage}{0.42\linewidth} 
    \begin{center}
    \begin{tabular}{>{\raggedleft}p{0.425\linewidth}
      >{\raggedleft}p{0.2\linewidth}      
      >{\raggedleft\arraybackslash}p{0.175\linewidth}}
    \toprule
     & Pooled & GGNN \\ 
     \midrule 
      F-score $> 0.5$ & 22.27\% & 28.37\% \\
      F-score $> 0.0$ & 29.82\% & 37.08\% \\
     \midrule 
     Entity linker errors & 6\% & 8\% \\
     No path to answer & 14\% & 18\% \\
     Data set inconsistency & 8\% & 14\% \\
     Wrong prediction & 72\% & 60\% \\
      \midrule     
  hit@10 & 35.62\% & 44.05\% \\     
    \bottomrule
    \end{tabular} \\
    \rule{0em}{0.25cm}
    \end{center}
    \captionof{table}{Manual error analysis results on WebQSP-WD \label{table:errors}}
\end{minipage} 
\end{figure}

\subsection{QALD-7} 

Next, we examine the out-of-domain results on the QALD-7 data set for the five models (see Table~\ref{table:qald-results}). The difference in performance is less prominent on this data set, but we can observe the same trends. The Single Edge model outperforms both the STAGG and Pooled Edges baselines. The GGNN delivers the best performance, although overall the best result is worse than on the WebQSP-WD data set. QALD-7 is much smaller, but also more complex on average (cf. the average number of edges needed to find the correct answer in Table~\ref{table:datasets}). Overall, we can conclude that the explicit modeling of the structure of semantic graphs with GGNN results in a better performance both in-domain on WebQSP-WD and out-of-domain on QALD-7 that was only used for evaluation.

\subsection{Error analysis}
\label{sec:error}

To better understand the difference between the approaches that encode graph structure and the approaches that disregard it, we closely look at the output of GGNN compared to the Pooled Edges model. The first two rows in Table~\ref{table:errors} show how often the respective model has returned an answer with F-score higher than $0.5$ which is a mostly correct answer and how often it returned an answer that was not just completely wrong (F-score $> 0.0$). We see that GGNN delivers an acceptable answer almost 25\% more often than Pooled Edges model, but there is still a lot of questions that are not answered correctly.

We have manually analyzed 100 sample answers from the two models where the resulting F-score was lower than $0.5$ (see rows 3 through 6 in  Table~\ref{table:errors}). Since GGNN makes less mistakes in general, the error propagation from the entity linking takes a slightly larger portion of the final error count. In 18\% of the cases, it was impossible to find the answer in Wikidata since there were no path between entities in the question and the answer. For example, for the question ``Where did Harper Lee attend high school?'', the correct answer ``Monroe County High School'' is a valid entity in Wikidata, but it is not connected to ``Harper Lee'' via the \textsc{educated at} relation. 14\% of the time, the data set contained an inconsistent answer and even though the model has predicted the correct semantic graph, the answers did not match. For example, a correct answer for a question about someone's place of birth is usually a city or a town, yet for a smaller set of questions a city borough (e.g. Manhattan) or a country are listed in the data set. 

Overall, in 32\% of the cases the error was caused by the gap between the KB and the data set. 
This lets us put the current results into a perspective with the previously reported numbers for the Freebase KB. If we approximately adjust our results for this kind of errors, we achieve between $0.469$ and $0.51$ F-score. \newcite{Reddy2016} report results for various approaches ranging from $0.404$ to $0.503$ F-score on the original WebQuestions data set that is a superset of WebQSP-WD (see Section~\ref{sec:datasets}).

The majority of errors for both models are caused by wrong predictions (row 6 in Table~\ref{table:errors}). GGNN selects significantly less wrong semantic graphs and more often successfully handles graphs with multiple edges. For example, for a question ``What language do people speak in Brazil?'', the GGNN model correctly predicts the graph with two edges \textsc{home country} and \textsc{native language} to get a list of all languages that are spoken in Brazil. Whereas the other models either select the relation \textsc{official language} that returns only the official language of the country or choose a wrong interpretation altogether.
We also look at the hit@10 measure that shows how often the correct semantic graph was in the top 10 scored graphs by the model (row 7 in Table~\ref{table:errors}). Notably, in 44\% of the cases the correct semantic graph was still among the top scored graphs for the GGNN model.

\section{Related work} 

We have focused on the problem of the increasing error rates for complex questions and the encoding of the semantic graph structure. 
In this paper, we describe a semantic parsing system for KB QA and follow the approach of \newcite{Yih2015} who do not rely on syntactic parsing to construct semantic parses. Our semantic graphs do not cover some aspects of the first-order logic, such as negation. 
\newcite{Reddy2016} define a semantic parser that builds first-order logic representations from syntactic dependencies. They further specify how it can be extended with negation in \mbox{\newcite{Fancellu2017}}.

We only train on the WebQSP-WD data set and we note that more data might be necessary to effectively train the gated graph architecture. \newcite{Reddy2014} suggest an unsupervised learning method to learn a model from a large web corpus, while \newcite{Su2016} use patterns and crowdsourcing to create new data with specific properties. These techniques can be used to further improve the performance of our model.

An alternative solution to semantic parsing is to build an information extraction pipeline that views the question as a query and the KB as a source of relevant information~\cite{Yao2014a}. \newcite{Dong2015} and \newcite{Jain} construct a vector representation for the question and use it to directly score candidate answers from the KB. However, such approaches are hard to analyze for errors or to modify with explicit constraints. For example, it is not directly possible to implement the temporal sorting constraint ($\mathrm{argmax}$).

We apply GGNNs to the problem of semantic parsing. \newcite{Li2015b} have developed the gated architecture based on the graph neural network formulation of \newcite{Scarselli}. Recently, a slightly different design of Graph Convolutional Networks  was proven effective on a KB completion task~\cite{Schlichtkrull2018}. \newcite{Kipf2017} introduced Graph Convolutional Networks, while \newcite{Marcheggiani2017} employed them for natural language processing for the first time and compared them to other formulations. Graph Convolutional Networks have a similar gated architecture and share most of the same properties with the Gated Graph Neural Networks used.

\section{Conclusions}
In this work, we have used Gated Graph Neural Networks to encode the structure of the target semantic parse for KB QA. We have shown that disregarding the semantic structure leads to a falling performance on questions that require complex semantic parses to get the correct answers. Our GGNN architecture was able to successfully model the structure of semantic parses. We have compared the performance of GGNNs against the previous work and non-graph models on two data sets and have broken down the results by question complexity. The analysis has shown that the suggested graph architectures do not have the same drop in performance on complex questions and produce better overall results.

Recently, \newcite{Peng2017} and \newcite{Yu2017} have attempted to incorporate entity linking into a feature based QA model. In the future, we plan to follow their work and  integrate GGNNs with entity linking into a single model. We also see possible applications for GGNNs on other semantic parsing tasks, such as text comprehension.

\section*{Acknowledgments}
This work has been supported by the German Research Foundation as part of the Research Training Group AIPHES (grant No. GRK 1994/1), and via the QA-EduInf project (grant GU 798/18-1 and RI 803/12-1). 
We gratefully acknowledge the support of NVIDIA Corporation with the donation of the Titan X GPU.

\bibliographystyle{acl}
\bibliography{qa-graph}

\end{document}